\documentclass[conference]{IEEEtran}
\IEEEoverridecommandlockouts
\usepackage{cite}
\usepackage{amsmath,amssymb,amsfonts}
\usepackage{physics}
\usepackage{algorithm}
\usepackage{algorithmic}
\usepackage{graphicx}
\usepackage{textcomp}
\usepackage{xcolor}
\usepackage{amsmath,bm}

\newcommand{\be}{\begin{equation}}
\newcommand{\ee}{\end{equation}}

\def\BibTeX{{\rm B\kern-.05em{\sc i\kern-.025em b}\kern-.08em
    T\kern-.1667em\lower.7ex\hbox{E}\kern-.125emX}}
\begin{document}

\title{User Scheduling for Federated Learning Through Over-the-Air Computation\\
}

\author{\IEEEauthorblockN{Xiang Ma\IEEEauthorrefmark{1}, Haijian Sun\IEEEauthorrefmark{2}, Qun Wang\IEEEauthorrefmark{1}, Rose Qingyang Hu\IEEEauthorrefmark{1}}
\IEEEauthorblockA{
\IEEEauthorrefmark{1}Department of Electrical and Computer Engineering, Utah State University, Logan, UT, USA \\
\IEEEauthorrefmark{2}Department of Computer Science, University of Wisconsin-Whitewater, Whitewater, WI, USA\\
Email: \IEEEauthorrefmark{1}\{xiang.ma@ieee.org, claudqunwang@ieee.org, rose.hu@usu.edu\}, \IEEEauthorrefmark{2}h.j.sun@ieee.org}

}

\maketitle

\begin{abstract}

A new machine learning (ML) technique termed as federated learning (FL) aims to preserve data at the edge devices and to only exchange ML model parameters in the learning process. FL not only reduces the communication needs but also helps to protect the local privacy. Although FL has these advantages, it can still experience  large communication latency when there are massive edge devices connected to the central parameter server (PS) and/or millions of model parameters involved in the learning process. Over-the-air computation (AirComp) is capable of computing while transmitting data by allowing multiple devices to send data simultaneously by using analog modulation. 
To achieve good performance in FL through AirComp, user scheduling plays a critical role.  In this paper, we investigate and compare different user scheduling policies, which are based on various criteria such as wireless channel conditions and the significance of model updates. Receiver beamforming is applied to minimize the mean-square-error (MSE) of the distortion of function aggregation result via AirComp. 
Simulation results show that scheduling based on the significance of model updates has smaller fluctuations in the training process while scheduling based on channel condition has the advantage on energy efficiency.

\end{abstract}


\section{Introduction}
The availability of big data makes data-driven artificial intelligent applications such as image recognition and autonomous driving ever increasingly realistic. Nowadays, advanced machine learning (ML) techniques usually comprise training and inference processes that work in a centralized manner. However, distributed devices such as smart sensors or unmanned aerial vehicles (UAVs)  have massive locally generated data and need to make real-time decisions, which render it extremely difficult to transmit data for central processing through wireless channels. Thanks to the rising capacity of computation, storage, and power at edge devices, they can perform ML tasks using locally collected raw data, which can largely reduce the communication overhead and latency. 

Although raw data is preserved and used locally and does not have to be uploaded to a central parameter server (PS), edge devices still need to coordinate with PS to establish the global model. A new machine learning technique named as federated learning (FL) appears to help address this issue \cite{federated_learning}. FL keeps the collected data locally and trains the ML model on edge devices. Only model parameters are transmitted to the PS for aggregation to attain the global model through averaging. There are usually a large number of edge devices connected to the PS and all the devices contend for limited wireless bandwidth. FL only selects a small subset of edge devices for model update in each communication round \cite{online_client,update_aware,scheduling}. Since the devices collect the data from their local environment, the data on different devices can be heterogeneous or non-i.i.d (independent and identically distributed). Thus it is important to select the most relevant devices for model update based on certain scheduling criteria in each round. In \cite{scheduling}, three scheduling policies, i.e., random scheduling, round robin, proportional fairness in terms of probability, group, and channel condition separately are proposed. It considered the channel conditions but neglected the data distribution on different devices.

To achieve spectrum efficiency, advanced transmission techniques can be used in model parameter uploading. Non-orthogonal multiple access (NOMA) \cite{noma} allows multiple devices to share the channel and transmit data simultaneously, which reduces the aggregation latency compared with the conventional time-based scheme. NOMA users use different transmit powers and successive interference cancellation (SIC) is applied at the PS side. The authors in \cite{adaptive} investigated the performance of FL under NOMA with 7x performance gain without loss of accuracy. However, it doesn't provide the security feature and the number of users that can transmit simultaneously is still limited due to the decoding at the receiver side. In \cite{over-the-air}, FL via over-the-air computation (AirComp) is presented. It employed the superposition nature of a wireless multiple-access channel to aggregate the model parameters while transmitting. PS deals with the aggregated model but not the individual, and does not have to decode the received signals like in nominal NOMA transmission. Therefore it is not only communication efficient but also computation efficient. Additionally, since PS cannot decode the received signal, it provides security features for FL as the dishonest PS cannot infer the local data with the aggregated model. 

The power control for AirComp in fading channels that minimizes the computation error is presented in \cite{power_allocation} \cite{optimal_power}. Authors in \cite{wireless_edge} evaluated the performance of AirComp in both digital approach and analog approach. In \cite{learning_rate}, the learning rate optimization of federated learning under AirComp is explored. However, no existing work has considered the user scheduling schemes for FL under AirComp that are significant to improve FL performance. 


In this work, we focus on FL via AirComp to improve both communication and computation efficiency. It employs the superposition nature of a wireless multiple-access channel so that multiple edge devices can transmit the model parameters simultaneously and PS does not need to decode the analog aggregated signals. To minimize the aggregated signal error, receiver beamforming design is applied. We  explore different scheduling schemes including channel based one, model update based one,  and a hybrid one based on both channel and model update.

The rest of the paper is organized as follows. Section II introduces the system model, AirComp scheme, and problem formulation. Section III presents several different user scheduling policies. Simulation results are shown in Section IV. Lastly, section V concludes the paper.

\section{System Model}
We consider an AirComp system with $M$ edge devices, each with a single antenna connected to the PS that is equipped with $N$ antennas. Multiple edge devices are allowed to transmit simultaneously on the same channel. The number of edge devices participating in the model update in each communication round is limited in order to minimize the distortion error and maximize  the testing model performance. Assume the maximum number of selected devices for transmitting in each round is $K$ under AirComp\cite{over-the-air}. The main notations used in the paper are summarized in Table \ref{Tab:notation}.

\begin{table}[h]
	\newcommand{\tabincell}[2]{\begin{tabular}{@{}#1@{}}#2\end{tabular}}
	\centering
	\caption{Summary of Notations}
	\begin{tabular}{c|p{65mm}}
		\hline
		\textbf{Notation} & \textbf{Definition}\\
		\hline
		M; K; W & The total number of edge devices connected to PS; the maximum number of edge devices participating FL in each round; the intermediate number of edge devices when considering both model update and channel condition\\
		\hline
		N; T; $S_K$ & The number of antennas at PS; the total number of communication round; Selected edge device set \\
		\hline
		$\mathbf{x}_k$; $\mathbf{y}_k$; $\bm{\theta}_k$; & Features of a data point sample on device $k$; corresponding label of data point; parameter set describe the mapping from $\mathbf{x}_k$ to $\mathbf{y}_k$ \\
		\hline
		$F(\cdot)$; $f(\cdot)$; $\eta$ & Global loss function; local loss function; learning rate \\
		\hline
        $\mathcal{D}_k$; $|\mathcal{D}_k|$ & Dataset on user $k$; cardinality of the dataset $\mathcal{D}_k$ \\
		\hline
		$\bm{h_k}$; $b_k$; $s_k$ & Channel vector of user $k$; transmitter scaling factor of user $k$; normalized local update at one time slot
		\\
		\hline
		$P_0$; $\phi_k(\cdot)$; $\psi(\cdot)$ & Maximum transmit power; pre-processing function of user $k$; post-processing function at PS\\
		\hline
		$\bm{r}$; $\bm{a}$; $\bm{n}$ & Received signal vector; receiver beamforming vector; additive noise \\
		\hline
		$g$; $\hat{g}$; $\tau$ & summation result before post-processing; estimation of $g$;
		normalizing factor\\
		\hline
	\end{tabular}
	\label{Tab:notation}
\end{table}

\subsection{FL System}
In FL, each edge device performs machine learning tasks using locally collected and stored data. For device $k$, data sample $\bm{x}_k$ has a label $\bm{y}_k$. Model parameters $\bm{\theta}_k$ is used to capture the mappings from $\bm{x}_k$ to $\bm{y}_k$. Each device executes stochastic gradient descent (SGD) updates to minimize the loss function that describes the loss of model parameter $\bm{\theta}_k$ at sample $\bm{x}_k$. The loss function at device $k$ is given by
\be
\setlength{\abovedisplayskip}{3pt}
\setlength{\belowdisplayskip}{3pt}
F_k(\bm{\theta}_k)=\frac{1}{|\mathcal{D}_k|}\sum_{\bm{x}_k \in{\mathcal{D}_k }}f(\mathbf{x}_{k}, \mathbf{y}_{k}; \bm{\theta}_k),
\ee
where $\mathcal{D}_k$ is the local dataset on device $k$, $|\mathcal{D}_k|$ is the cardinality of $\mathcal{D}_k$, $f(\mathbf{x}_{k}, \mathbf{y}_{k}; \bm{\theta}_k)$ is the empirical loss function. The entire empirical loss function across dataset $\{ \mathcal{D}_1, \mathcal{D}_2, \ldots, \mathcal{D}_K\}$ can be written as

\be
\setlength{\abovedisplayskip}{6pt}
\setlength{\belowdisplayskip}{3pt}
F(\bm{\theta})= \sum_{k=1}^K \frac{|\mathcal{D}_k |}{|\mathcal{D}|}  F_k (\bm{\theta}_k),
\ee
where $|\mathcal{D}| = \sum_{k=1}^K |\mathcal{D}_k|$,  $\bm{\theta}$ is the global model parameters by averaging the aggregation result
\be
\setlength{\abovedisplayskip}{3pt}
\setlength{\belowdisplayskip}{3pt}
\bm{\theta}(t+1)= \frac{1}{K}\sum_{k=1}^K \bm{\theta}_k(t+1).
\ee

To reduce the communication overhead, the local model update $\Delta \bm{\theta}_k(t+1)$ rather than the local model $\bm{\theta}_k(t+1)$ itself is uploaded. Thus the aggregation result can be written as 
\be
\setlength{\abovedisplayskip}{3pt}
\setlength{\belowdisplayskip}{3pt}
\bm{\theta}(t+1)= \bm{\theta}(t) + \frac{1}{K}\sum_{k=1}^K \Delta \bm{\theta}_k(t+1),
\ee
where $\Delta \bm{\theta}_k(t+1) \triangleq \bm{\theta}_k(t+1) - \bm{\theta}_k(t)$, is defined as the local model update at devices $k$.

\begin{figure}[!h]
    \centering
	\includegraphics[width=45mm, keepaspectratio]{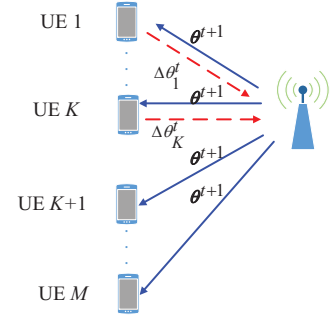}
	\caption{FL Model Update}
	\label{fig:system_model}
\end{figure}

Fig. \ref{fig:system_model} shows the FL update model.
\subsection{AirComp Scheme}
AirComp performs transmission and computation simultaneously over the air. Unlike traditional orthogonal multiple access schemes, AirComp allows multiple transmission via the same channel simultaneously. It performs analog modulation and waveform superposition and no individual decoding is needed at the receiver side.  Since AirComp does not decode the signal at the PS side, PS does not know the model parameters of the individual user. Thus it cannot infer local data information of individual users, providing a more secured transmission scheme.  

Since the aggregation takes place during over-the-air transmission,  the received signal at PS is given by
\be
\setlength{\abovedisplayskip}{3pt}
\setlength{\belowdisplayskip}{3pt}
\bm{r} = \sum_{k=1}^K\bm{h}_kb_ks_k+\bm{n},
\ee
where $\bm{h_k}$ is the channel vector between device $k$ and PS, $b_k$ is the transmitter scaling factor, $s_k$ is normalized local update $\bm{s_k}$ at one time slot, where $\bm{s_k} \triangleq \bm{\theta}_k$ with unit variance, i.e. $||\bm{s}_k||_2^2 = I$, $\bm{n} \sim \mathcal{CN} (0, \sigma^2\bm{I})$ is the noise vector. The transmit power constraint at device $k$ is $
E(|b_ks_k|^2)=|b_k|^2 \leq P_0$, 
where $P_0$ is the maximum transmit power.

The target function at the PS side that is computable over-the-air can be written as $
v = \psi(\sum_{k=1}^K\phi_k(s_k))$,
where $\phi_k (x) =|\mathcal{D}_k| x$ is the pre-processing function of user $k$, and $\psi (x) =\frac{1}{|\mathcal{D}|} x$ is the post-processing function at PS side. The weighted summation of transmitted signal is 
\be
\setlength{\abovedisplayskip}{3pt}
\setlength{\belowdisplayskip}{3pt}
g = \sum_{k=1}^K\phi_k(s_k).
\ee
The received signal at PS is $\bm{r}$. Then the estimated value at after beamforming is
\be
\setlength{\abovedisplayskip}{3pt}
\setlength{\belowdisplayskip}{3pt}
\hat{g}= \frac{1}{\sqrt{\tau}}\bm{a}^{\mathsf{H}}\bm{r} = \frac{1}{\sqrt{\tau}}\bm{a}^{\mathsf{H}}\sum_{k=1}^K\bm{h}_kb_ks_k+\frac{\bm{a}^{\mathsf{H}}\bm{n}}{\sqrt{\tau}},
\ee
where $\tau$ is the normalizing factor and $\bm{a}$ is the receiver beamforming vector. The distortion of $\hat{g}$ with respect to the target value $g$, which quantifies the AirComp performance,  is measured by the mean-square-error (MSE) given by
\be
\begin{aligned} \label{eq:mse_def}
\setlength{\abovedisplayskip}{3pt}
\setlength{\belowdisplayskip}{3pt}
{\sf MSE}(\hat{g},g)&=E(|\hat{g}-g|^2)\\&=\sum_{k=1}^K|\frac{1}{\sqrt{\tau}}\bm{a}^{\mathsf{H}}\bm{h}_kb_k-\phi_k|^2+\frac{\sigma^2||\bm{a}||^2}{\tau}.
\end{aligned}
\ee

We choose parameters $b_k$ and $\bm{a}$  to minimize the MSE. Supposing the receiver beamforming vector $\bm{a}$ is given, the transmitter scaling factor $b_k$ can be selected  by using a uniform-forcing transmitter as \cite{uniform_forcing}
\be
\setlength{\abovedisplayskip}{3pt}
\setlength{\belowdisplayskip}{3pt}
b_k=\sqrt{\tau} \phi_k\frac{(\bm{a}^{\mathsf{H}}\bm{h}_k)^{\mathsf{H}}}{||\bm{a}^{\mathsf{H}}\bm{h}_k||^2}.
\ee
The normalizing factor $\tau$ can be calculated as 
\be
\setlength{\abovedisplayskip}{3pt}
\setlength{\belowdisplayskip}{3pt}
\tau = P_0 \min_k \frac{||\bm{a}^{\mathsf{H}}\bm{h}_k||^2}{\phi_k^2}.
\ee
Then the corresponding MES problem can be calculated as 
\be \label{eq:mse}
\setlength{\abovedisplayskip}{3pt}
\setlength{\belowdisplayskip}{3pt}
{\sf MSE} = \frac{||\bm{a}^{\mathsf{H}}||^2\sigma^2}{\tau} = \frac{\sigma^2}{P_0}\max_k \frac{\phi_k^2||\bm{a}^{\mathsf{H}}||^2}{||\bm{a}^{\mathsf{H}}\bm{h}_k||^2}.
\ee

To achieve the best performance, the following minimum mean square error is applied. 
\be
\setlength{\abovedisplayskip}{3pt}
\setlength{\belowdisplayskip}{3pt}
\min_{\bm{a}} \max_k \frac{\phi_k^2||\bm{a}^{\mathsf{H}}||^2}{||\bm{a}^{\mathsf{H}}\bm{h}_k||^2}.
\ee

It can be formulated into a more friendly way as 
\be \label{eq:qcqp}
\setlength{\abovedisplayskip}{3pt}
\setlength{\belowdisplayskip}{3pt}
\begin{aligned}
\min_{\bm{a}} \quad & ||\bm{a}||^2 \\
\textrm{s.t.} \quad & \frac{||\bm{a}^{\mathsf{H}}\bm{h}_k||^2}{\phi_k^2} \geq 1.
\end{aligned}
\ee 

Eq. (\ref{eq:qcqp}) is a quadratically constrained quadratic programming (QCQP) problem with non-convex constraints, which is still hard to solve. In \cite{uniform_forcing}, the same problem can be solved by semidefinite programming (SDP), improved by successive convex approximation (SCA).  After the receiver vector $\bm{a}$ is solved, all other parameters can also be calculated. And the minimum MSE can be obtained.

 \textbf{Algorithm 1} summarizes the SDP and SCA method to optimize the receiver vector.

\begin{algorithm}[]
\caption{Receiver Optimization by SDP and SCA}
\begin{algorithmic}[1]
\STATE  SDP method to obtain $\bm{A}^*$
\IF{rank($\bm{A}^* \neq 1$)}
\STATE $\bm{\tilde{a}^*}=\sqrt{\lambda_1}\bm{u_1}$
\STATE Set $\bm{c_k}=[\Re({\bm{\tilde{a}}}^{*\mathsf{H}} \bm{h}_k), \Im({\bm{\tilde{a}}}^{*\mathsf{H}} \bm{h}_k)], \forall k$
\REPEAT 
\STATE SCA method solve $\frac{||\bm{c_k}||^2}{\phi_k^2} \geq 1$ to obtain $\bm{a}$ and $\bm{c_k}$
\UNTIL criteria satisfied
\ELSE
\STATE $\bm{a}=\sqrt{\lambda_1}\bm{u_1}$
\ENDIF
\end{algorithmic}
\end{algorithm}

Here, $\bm{A}^*=\min_{\bm{A}}\tr(\bm{A})$ and , $\bm{A}=\bm{a} * \bm{a}^H$, $\lambda_1$ is the largest eigenvalue of $\bm{A}^*$ and $\bm{u_1}$ is the corresponding eigenvector. $\bm{c_k}$ is the auxiliary variable.

\textbf{Algorithm 2} summarizes the  proposed FL process under AirComp settings.

\begin{algorithm}[]
\caption{FL in AirComp}
\begin{algorithmic}[2]
\STATE  {\bf Initialization:} $\bm{\theta}^0$, $T$. 
\FOR {each FL update round $t$} 
\STATE PS sends $\bm{\theta}^t$ to all users
\FOR {each user $i$ in parallel} 
\STATE {Calculate local gradients:   $\bm{\theta}_i^{t} =  \bm{\theta}_i^{t} - \eta \nabla F_i(\bm{\theta})$. }
\ENDFOR
\STATE PS selects $K$ users based on scheduling algorithm. 
\STATE Selected users send gradients $\nabla F_k(\bm{\theta})$ to the PS simultaneously via AirComp. 
\STATE PS samples the received signal to get aggregated model. 
\ENDFOR
\end{algorithmic}
\end{algorithm}

\section{User Scheduling Policies}
There are usually a large number of edge devices connected to the PS. Although AirComp allows multiple users to upload their model simultaneously, the maximum number of users participating in model update in each round is normally still smaller than the total number of users \cite{over-the-air}. Here, we consider an FL system with a total of $M$ devices connected to the PS while  $K$ devices can be scheduled in each round, $K < M$. We propose three user scheduling policies, one considers  channel conditions from communication perspective, one considers the significance of local model update from computation perspective, and one considering both. Correspondingly the three scheduling policies are named channel based scheduling,  model update based scheduling, and hybrid scheduling.
\subsection{Channel Based Scheduling}
Channel based scheduling selects $K$ users that have the highest channel gains, i.e.,  
\be
\setlength{\abovedisplayskip}{3pt}
\setlength{\belowdisplayskip}{3pt}
S_K = \max_{[K]}\{||\bm{h}_1(t)||, \dots, ||\bm{h}_M(t)||\}.
\ee
here, $||\bm{h}_k(t)||=\sqrt{\sum_{i=1}^N|h_k^i(t)|^2}$ is the $l_2$-norm channel gain of device $k$. Before scheduling, each client needs to send a small amount of information to PS so that the  PS can perform channel estimation. Compared with the model gradient transmission, the time to transmit this small amount of information can be safely ignored.

Since multiple antennas are equipped in PS, channel gain is in a vector form. From Eq. \eqref{eq:mse},  a larger channel gain results in a smaller MSE when other parameters are fixed.

In this scheduling scheme, users can start local computation until they are selected. Thus, energy-constrained edge devices such as IoT devices can be more power efficient.
\subsection{Model Update Based Scheduling}
This scheduling scheme considers the significance of the model update as the user selection criteria. $l_2$-norm is used to evaluate the significance of model update. Edge device $k$, $k=1,...,M$,  first computes the model update $\Delta \bm{\theta}_k(t)$ and then sends its $l_2$-norm of model update $||\Delta \bm{\theta}_k(t)||_2$ to the PS. Then PS selects $K$ devices with the largest $||\Delta \bm{\theta}_k(t)||_2$ value, that is 
\be
\setlength{\abovedisplayskip}{3pt}
\setlength{\belowdisplayskip}{3pt}
S_K = \max_{[K]}\{||\bm{\theta}_1(t)||, \dots, ||\bm{\theta}_M(t)||\}.
\ee
This scheme requires all the users to perform local computation and send their $l_2$-norm of model update to the PS. It causes energy dissipation for the unselected devices and the transmission of model update for all the users can also cause channel congestion. For devices with low computation abilities, it may take a long time for them to finish the local computation and upload their model update. The stragglers will reduce system performance.
\subsection{Hybrid Scheduling}
Channel gain and the significance of model update can both affect the performance of FL. Thus both are considered in the hybrid scheduling.  PS first selects $W$ devices with the highest channel gains and then selects $K$ devices with the largest model update from  $W$ devices, $K \leq W \leq M$. In this strategy, the energy of unselected devices can be saved since only selected devices need to perform local computation.

Channel based scheduling can help reduce computation needs at local devices while  model update based scheduling can help improve the FL training performance. Hybrid scheduling intends to balance the tradeoff between the two. 

\subsection{Complexity Analysis}
For each client, supposing the computation time to finish the ML task is $t_p$, the communication time for PS channel estimation is $t_o$ and the communication time to upload model gradients is $t_u$. The corresponding time complexity is summarized in Table \ref{Tab:complexity}.

\begin{table}[h]
	\newcommand{\tabincell}[3]{\begin{tabular}{@{}#1@{}}#2\end{tabular}}
	\centering
	\caption{Complexity Analysis\label{Tab:complexity}}
	\begin{tabular}{p{17mm}|p{17mm}|p{21mm} | p{13mm}}
		\hline
		 & Channel Based Scheduling &
		 Model update Based Scheduling & Hybrid Scheduling \\
		\hline
		Communication  & $M*t_o+K*t_u$ & $K*(t_o+t_u)$ & $M*t_o+K*t_u$ \\
		\hline
		Computation  & $K*t_p$ & $M*t_p$ & $W*t_p$ \\
		\hline
		
	\end{tabular}
\end{table}

\section{Simulation Results}
In this section, we present the performance of federated learning under AirComp with different user scheduling schemes. The channel parameters are given as follows. There are $M=1000$ users uniformly distributed in a disk region with a  cell size of $500$ m. The transmit signal to noise ratio $\frac{P0}{\sigma^2}$ is fixed at $42$ dB, channel path loss exponent is $\alpha=3$. The number of antennas at PS is $N=4$. In each communication round, the channel vector keeps constant for the same user while it varies across different users and/or different communication rounds. We further have  $K=10$ and $W=20$. The learning task is trained by using the MNIST (Modified National Institute of Standards and Technology) dataset \cite{mnist} with a fully connected neural network called LeNet-300-100, where the first hidden layer consists of $300$ neurons and the second layer  consists of  $100$ neurons. The hyperparameters are summarized in Table \ref{Tab:hyper}. The learning stages are divided into two phases, namely training phase and testing phase. Similarly, the dataset also split into two parts, $90\%$  of them are training set and the rest are the testing set. The testing accuracy is used to evaluate the learning performance. To make the proposed scheduling schemes more convincing, non-i.i.d data \cite{fedprox} is used here, i.e., every user has a varying data size and distribution. 

\vspace{-0.1in}
\begin{table}[h]
	\newcommand{\tabincell}[2]{\begin{tabular}{@{}#1@{}}#2\end{tabular}}
	\centering
	\caption{Hyperparameters\label{Tab:hyper}}
	\begin{tabular}{p{14mm}|p{13mm}|p{13mm}|p{12mm} | p{12mm}}
		\hline
		\textbf{Learning \newline rate size ($\eta$)} & \textbf{Batch \newline size ($\mathcal{B}$)} &
		\textbf{FL \newline Round ($T$)} & \textbf{Training \newline set size} & \textbf{Testing \newline set size}\\
		\hline
		0.01 & 10 & 60 & 90\% & 10\%\\
		\hline
	\end{tabular}
\end{table}

\vspace{-0.25in}
\begin{figure}[!h]
    \centering
	\includegraphics[width=60mm]{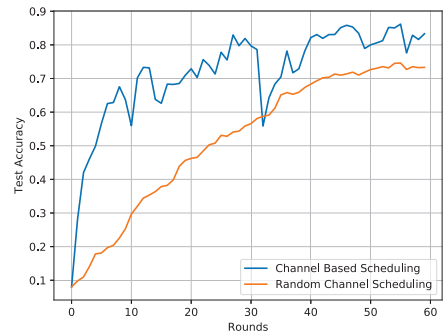}
	\caption{Channel Based Scheduling}
	\label{fig:channel_based}
\end{figure}


Fig. \ref{fig:channel_based} shows the testing accuracy for the channel based scheduling. The random channel scheduling  selects the client with different channel conditions in a uniform distribution. Compared with random channel scheduling, channel based scheduling achieves a much higher testing accuracy during the updating process but it experiences much larger fluctuations. This is because the MSE (defined in equation \eqref{eq:mse_def}) achieved by the channel based scheduling is much smaller than that achieved by the  random scheduling and non-i.i.d data causes the testing accuracy to drop in some rounds due to the inconsistency of the data updated. For random channel scheduling, the testing accuracy experiences smaller fluctuations because the impact from channel conditions plays down the impact from the non-i.i.d data distribution. 

\vspace{-0.15in}
\begin{figure}[!h]
    \centering
	\includegraphics[width=60mm]{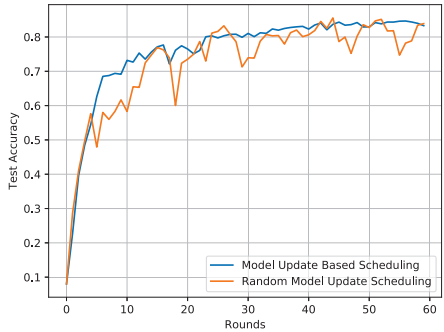}
	\caption{Model Update Based Scheduling}
	\label{fig:model_based}
\end{figure}

Fig. \ref{fig:model_based} gives the testing results when users with the largest model update values are selected. Compared with the scheduling that randomly selects model updates, the testing results of scheduling that selects the largest model updates are much smoother and the testing results of random scheduling are quite close to the model update based scheduling. In FL the gradients rather than the model parameters are uploaded here. As most of the gradient values are  close to $0$, there is no much difference between model update based scheduling and random scheduling \cite{sparsification}.

\vspace{-0.15in}
\begin{figure}[!h]
    \centering
	\includegraphics[width=58mm]{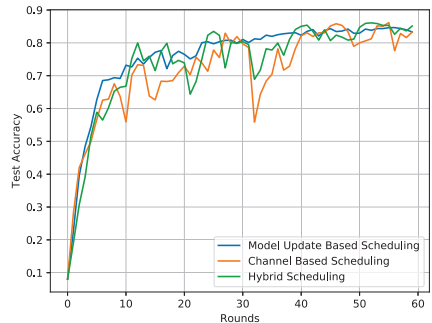}
	\caption{Hybrid Scheduling}
	\label{fig:both_considered}
\end{figure}

In Fig. \ref{fig:both_considered}, three scheduling schemes are compared. The model update based scheduling makes the testing result smoother while the channel based scheduling shows the lowest testing accuracy. The hybrid scheduling achieves the performance that falls in between the above two schemes.  In model update based scheduling,  all edge devices perform local computing, and devices with the largest model update values are scheduled for uploading. Thus the dataset is not non-i.i.d and the testing accuracy curve is quite smooth. However, the model update based scheduling consumes more computations  than the other two scheduling policies since all simulated edge devices need to perform local computing for the ML tasks. And local devices tend to consume energy for computing. Hybrid scheduling gives a good trade-off between the testing accuracy performance and local device energy consumption.

\section{Conclusion}
 AirComp based FL is not only communication efficient by allowing multiple devices to transmit simultaneously but also computation efficient since FL server only needs to have aggregated model parameters rather than the individual model parameter. To further investigate the AirComp based FL performance, in this paper, we proposed three different user scheduling policies, i.e., channel based scheduling, the model update based scheduling, and  a hybrid scheduling that consider both channel conditions and model update priorities. Simulation results show that the channel based scheduling has the least device computation needs but gives the lowest testing accuracy performance while the model update based scheduling gives the best testing accuracy result but has the highest computation needs. The hybrid scheduling basically gives the performance trade-off between the two. 

\section*{Acknowledgment}
This work was supported by the National Science Foundation under the grants NSF CNS-2007995 and EEC-1941524.

\end{document}